\documentclass[conference]{IEEEtran}
\IEEEoverridecommandlockouts
\usepackage[utf8]{inputenc}
\usepackage{cite}
\usepackage{amsmath,amssymb,amsfonts}
\usepackage{algorithmic}
\usepackage{graphicx}
\usepackage{textcomp}
\usepackage{xcolor}
\usepackage{array}
\usepackage{booktabs}
\usepackage{todonotes}
\usepackage{subfig}
\usepackage{multicol}
\graphicspath{{img/}}

\begin{document}

\title{Quantifying the presence of graffiti in urban environments\\
%TODO: PUT IT LATER
%{
%\thanks{}
}

%\author{\IEEEauthorblockN{Undisclosed authors}}
\author{\IEEEauthorblockN{Eric K. Tokuda, Roberto M. Cesar-Jr.}
\IEEEauthorblockA{\textit{Institute of Mathematics and Statistics} \\
\textit{University of São Paulo}\\
\textit{SP, Brazil} \\
\textit{\{tokudaek,rmcesar\}@usp.br}}
\and
\IEEEauthorblockN{Claudio T. Silva}
\IEEEauthorblockA{\textit{Tandon School of Engineering} \\
\textit{New York University}\\
\textit{NY, USA}\\
csilva@nyu.edu}
}

\maketitle

\begin{abstract}

Graffiti is a common phenomenon in urban scenarios. Differently from urban art, graffiti tagging is a vandalism act and many local governments are putting great effort to combat it. The graffiti map of a region can be a very useful resource because it may allow one to potentially combat vandalism in locations with high level of graffiti and also to cleanup saturated regions to discourage future acts. There is currently no automatic way of obtaining a graffiti map of a region and it is obtained by manual inspection by the police or by popular participation. In this sense, we describe an ongoing work where we propose an automatic way of obtaining a graffiti map of a neighbourhood. It consists of the systematic collection of street view images followed by the identification of graffiti tags in the collected dataset and finally, in the calculation of the proposed \emph{graffiti level} of that location. We validate the proposed method by  evaluating the geographical distribution of graffiti in a city known to have high concentration of graffiti -- São Paulo, Brazil.

\end{abstract}
\section{Introduction}

\begin{figure}[ht]
\centering
	\includegraphics[width=0.4\textwidth]{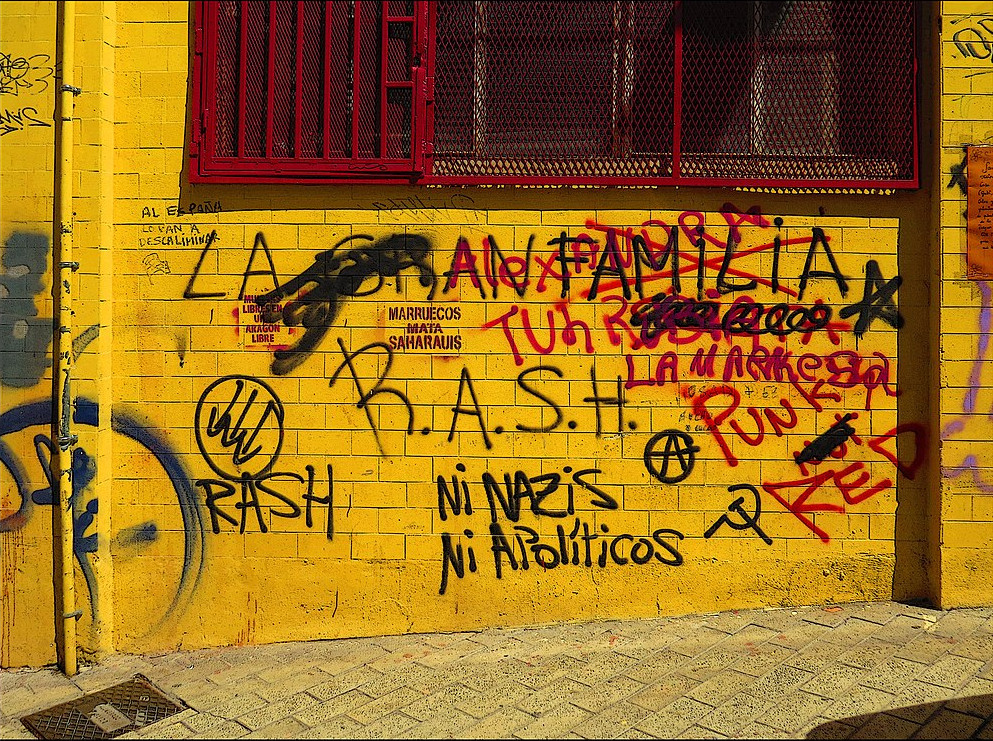} 
	\caption{The graffiti we interested in is composed of words and tags~\cite{googlemaps}.}
\label{fig:grafftypes}
\end{figure}

Graffiti is already part of the current landscape of most megacities. It can be categorized as artistic drawings or tagging and  whereas graffiti drawing is an artistic expression and as such, requires talent and practice, most of the times graffiti tagging represents an unauthorized act that people convey messages or show their names (see Figure~\ref{fig:grafftypes}). The discussion whether graffiti is an art is extensively explored~\cite{mcauliffe2012graffiti,young2013street}. In 2017 a Brooklyn based company was fined 6.7 million dollars for whitewashing the murals containing famous graffiti~\footnote{https://www.nytimes.com/2018/02/12/nyregion/5pointz-graffiti-judgment.html}. In this study we focus on the latter and refer to it as simply graffiti. 
  %Studies reveal that there is a strong correlation of people that perform these acts and future criminals~\cite{taylor2012patterns}.
%and major cities combat these acts.

%São Paulo is big city with an estimated GDP of $U\$580$ billion being the richest and a most influential city in South America. There is a high variability in ethnicity and with social levels, which contribute to form a multi-cultural city. Graffiti tagging is spread throughout the city~\cite{theguardian2016pixacao}.

There is no São Paulo map of graffiti to our knowledge and creating it by manual inspection would demand great effort. In this work we propose the creation of a graffiti map based on the segmentation of graffiti regions on street view images~\cite{googlemaps}.

Our contribution can be summarized as the proposal of an automated way to quantify the level of graffiti in a location. For this task, street view images are systematically acquired, computer vision algorithms identify and quantify the amount of graffiti in a picture and finally a new metric, the \emph{graffiti level} of the region is devised and computed. We perform a case study in a highly urbanized city, São Paulo, Brazil

% ##########################################################
\section{Related Work}

%There are many initiatives on the combat of graffiti.
%and also several works on image segmentation algorithms and on the use of street view imagery for urban analysis.

%We differ urban art by graffiti tags (see Figure~\ref{fig:grafftypes}).
%  “City, State Abbreviation.” Map. Google Maps. Google, 15 May 2008. Web. 15 May 2008.  

% ##################### Related work
% Patterns of graffiti offending: Towards recognition that graffiti offending is more than ‘kids messing around
%Graffiti writing is primarily an adolescent action and as such, has always been seen as a low-level crime ~\cite{taylor2012patterns}. However, according to~\cite{taylor2012patterns} there is a trend of felonies commited by graffiti writers, not only in terms of types of crimes commited, but also in the frequency of the crimes.

%Cost
%The Australian government estimates a total cost of \$833 million Australian dollars to repair vandalism and graffiti acts in 2011 ~\cite{smith2014counting}.

% Graffiti as art

%There are many initiatives on the combat of graffiti.
There are several works aimed to combat graffiti vandalism acts. Web-based frameworks~\cite{graffititracker,tagrs,graffititracking,vandaltrack} rely on community participation to identify recently degraded locations while some works~\cite{angiati2005novel,di2008graffiti, tombari2008graffiti} try to identify the drawing act. Other works retrieve similar graffitis in a reference database~\cite{yang2012efficient} by using connected components and keypoint matching in an attempt to associate graffiti with gangs. Alternatively, ~\cite{tong2011gang} attempts to identify the authorship, given a target image, they compute a metric based on the symbols contained, manually annotate and do a keypoint matching between the images and the gang graffitis.

%graffolution
As a signal of the relevance of the topic, the European Union has a dedicated project~\cite{synyo2016graffolution} to analyze the main actors involved in the graffiti acts, including writers, citizens, law enforcement and public administration. It also includes the consultation of stakeholders and the establishing of a web-based platform that allows the discussion and sharing of ideas about the topic from different perspectives. %A number of places alternatively apply more radical measures against these acts~\cite{iveson2010wars}.

% In Brazil, highly densed cities such as São Paulo also face the widespread presence of graffiti in the city~\cite{theguardian2016pixacao}.

%Image retrieval.
%Image retrieval using Sift~\cite{jain2009graffiti}.
%Identify in a fast way spatial and temporal change to identify the act of vandalism using multiple cameras~\cite{di2008graffiti, tombari2008graffiti}

%São Paulo is the richest and a very influential city in Brazil~\cite{ibge2016produto}. The city presents high social unequality which contributes to the widespread presence of graffiti~\cite{theguardian2016pixacao}.

Semantic segmentation is a high-level computer vision task which aims to split the image into known classes~\cite{arbelaez2012semantic}. It is a complex task when compared to image classification and object detection because it requires the image pixel-wise classification. Research on semantic segmentation is very active and recent works achieve impressive results~\cite{arbelaez2012semantic,ronneberger2015u,badrinarayanan2015segnet,zhou2017scene,long2015fully}. A related task is the \emph{instance segmentation}, where the objective is also to identify the instances but in contrast to the ordinary segmentation, the method must be able to identify the boundaries of two adjacent instances. Some previous works~\cite{pinheiro2015learning,dai2016instance} performed this task by preceding the object detection stage with a segmentation stage. The work of Mask-RCNN~\cite{he2017mask} in turn performed this task by running the classification and the segment proposals in a parallel manner. \cite{he2017mask} relies on Faster-RCNN~\cite{ren2015faster} architecture but with an additional branch for instance segmentation.

Google Maps\cite{googlemaps} provides public access to images captured by cars driven down the streets. Images are obtained from different geographical locations and in different views and formats. Many works~\cite{rundle2011using,torii2009google} have already utilized this type of imagery for urban analyzes.  The authors of~\cite{doersch2012makes} use street view images to compare the architectural elements from different cities. In~\cite{rundle2011using}, a study on the feasibility on audits of neighbourhoods environments instead of in-person auditing is presented. The work~\cite{li2015assessing} proposes the assessment of urban greenery using the same kind of images.

\section{Materials and Methods}

%The description of this work here presented is organized in two sections: image acquisition and graffiti recognition.
In order to confidently estimate the level of graffiti in a geographical region we propose a metric, the \emph{graffiti level}, obtained through the identification and computation of the areas containing graffiti on street view imagery~\cite{googlemaps}.

%###############################
\subsection{Acquisition}
% We use mask rcnn but we do not store the instance segmentation
The region of interest is initially defined and the images are acquired. Due to limitations on the coverage of the pictures provided and on computational constraints, a sample of the full region is considered. There are a number of ways of performing sampling~\cite{stehman1999basic} but they can be classified in random sampling methods and systematic sampling methods. The first removes the bias of the selection by randomly selecting the sample points although not guaranteeing good coverage. The latter, in contrast, assures coverage by including bias.

\begin{figure}[h]
        \centering
        \includegraphics[width=0.11\textwidth]{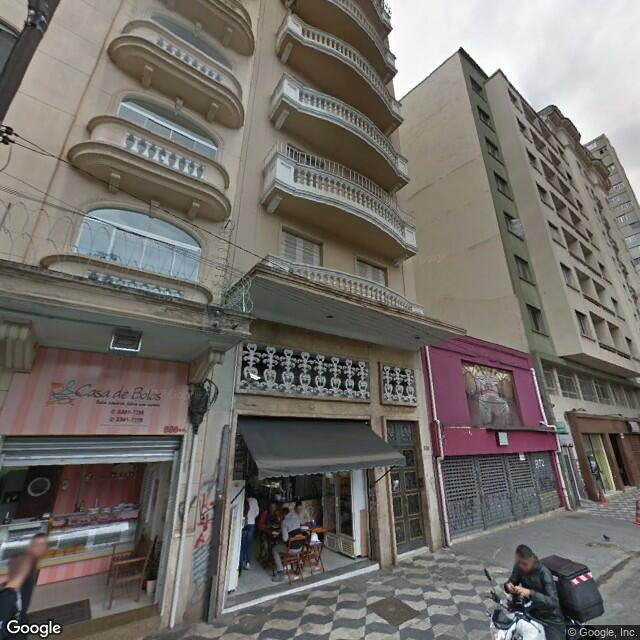} 
        \includegraphics[width=0.11\textwidth]{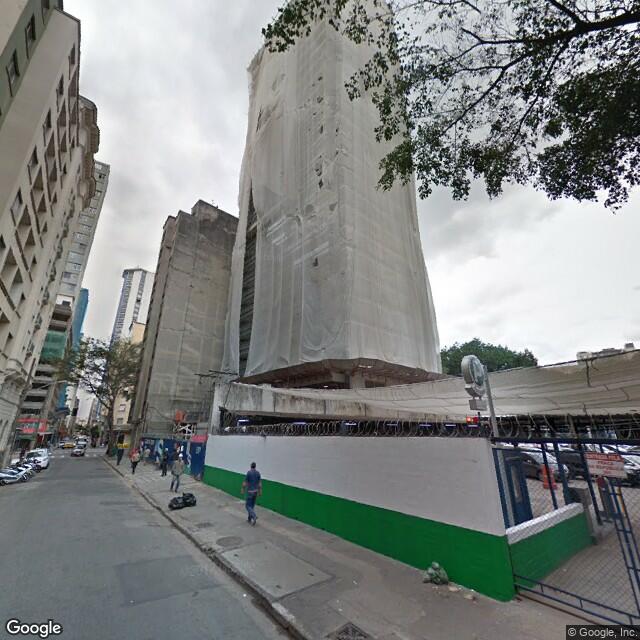} 
        \includegraphics[width=0.11\textwidth]{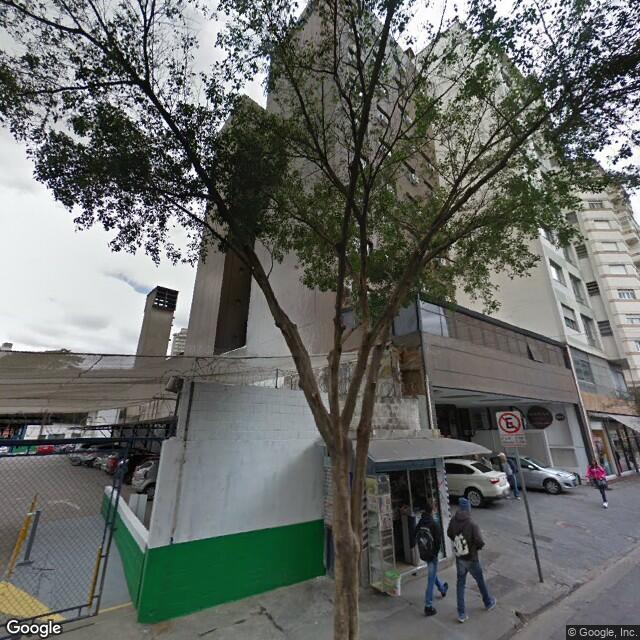} 
        \includegraphics[width=0.11\textwidth]{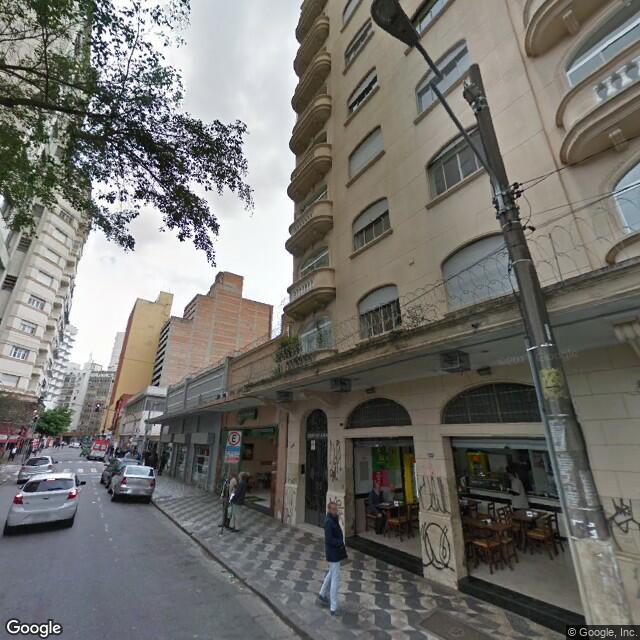} 
	\caption{Four views from the same geographical location. Images obtained from~\cite{googlemaps}.}
        \label{fig:views}
\end{figure}

Once the geographical sample is defined, ideally a full view should be considered for each geographical location. A single $360^{\circ}$ panorama view can be used, but one may need to worry about the distortions present in panoramic photos. Alternatively, complementary views for each location may be considered (see Figure~\ref{fig:views}).

%Discussion on the metric to evalute the level of degradation of a region
%We do not provide google images, but we do provide all the information required to download it from GSV.
 %* How to sample a geographical region? pros/cons
 %* How to handle multiple orientations of the camera? pros/cons

%###############################
\subsection{Graffiti recognition}

Given the objective of quantifying the level of graffiti in a given location, a simple and direct way would be to identify the images containing or not graffiti. Such characterization however give us a sharp and non-precise information of each picture. It would be interesting to have a more granular value for each picture. So we define the graffiti level $f(P)$ of a geographical location $P$ as the total area $A$ of the picture containing graffiti. Such approach can be affected by the projection map of the scene and also by the distance of the camera to the region containing graffiti. We assume that different regions, given a minimum extent, have corresponding distribution of projections and of distance to the walls and with this assumption, $f(P)$ may be used to compare different geographical locations.  Since we represent each location $P$ by a set of $k$ views, we define $f(P)$ as the sum of the areas of the regions containing graffiti in each view (see~(\ref{eq:perphoto})). We can then aggregate the graffiti level by geographical regions $R$ by computing the average of the graffiti levels on our sample of size $n$ (see~(\ref{eq:perregion})) .

%Since we represent each point by a set of images each representing a view, we .

%\begin{equation}
%	G(P) = \sum_{i=1}^{k} A_i
%	\label{eq:perphoto}
%\end{equation}
%
%\begin{equation}
%	G(R) = \frac{\sum_{j=1}^{n} G(P_j) }{n}
%	\label{eq:perregion}
%\end{equation}

\noindent\begin{minipage}{.34\linewidth}
  \begin{equation}
	f(P) = \sum_{i=1}^{k} A_i
	\label{eq:perphoto}
  \end{equation}
  \end{minipage} \qquad %
\begin{minipage}{.5\linewidth}
  \begin{equation}
	g(R) = \frac{\sum_{j=1}^{n} f(P_j) }{n}
	\label{eq:perregion}
  \end{equation}
  \end{minipage}
  \vspace{0.5cm}

%We opted for Mask-RCNN~\cite{he2017mask} method for our segmentation task given its high performance reported on important benchmarks~\cite{cordts2016cityscapes,lin2014microsoft}. Despite it outputs segmentation and instances information, in this work we are just interested on segmentation results.
We opted for Mask-RCNN~\cite{he2017mask} method for our segmentation task given its high performance reported on important benchmarks~\cite{cordts2016cityscapes,lin2014microsoft}. During training, the method minimizes a multi-task loss $L=L_{cls}+L_{box}+L_{mask}$, being the classification loss $L_{cls}$  and bounding-box loss $L_{box}$ the same as defined in~\cite{ren2015faster} and the mask loss $L_{mask}$ defined as the average binary cross-entropy loss. 

Since there is no dataset publicly available, we created a dataset with manually annotated images which were used to train our model.

 %* Any scratch on the wall is a graffiti?
 %* Why did I choose mask rcnn?
 %* Is it possible to train a graffiti detector without a training set?
 %* What is image augmentation
 %* How to measure the level of graffiti in an image? pros/cons

%###############################
%\subsection{Case study: São paulo}
%\subsection{Evaluation}
 %* Why Sao Paulo?
 
%We evaluated 
%The most populous city in Latin America.

%Social and economical importance in the country.

%Famous graffiti spots.

% wikimedia commons 
% wall colorful graffiti tags - jwh
% child and guard - Pawel Ryszawa,

% chicao- tags - John H. White, ]
% graffiti wall rose - no author provided
% graffiti wall yellow - agracier

% TODO: Include figure with annotation im_-23.541283999..._-46.6359..._280.jp
%im_-23.54178..._-46.63592...280.jpg

% ##################### Experiments
\section{Experiments}

%############################## Acquisition
%Google Street View api
% https://www.google.com/permissions/geoguidelines/attr-guide.html
%TODO: CITE above!

\begin{figure}[h]
	\centering
	\includegraphics[width=0.45\textwidth]{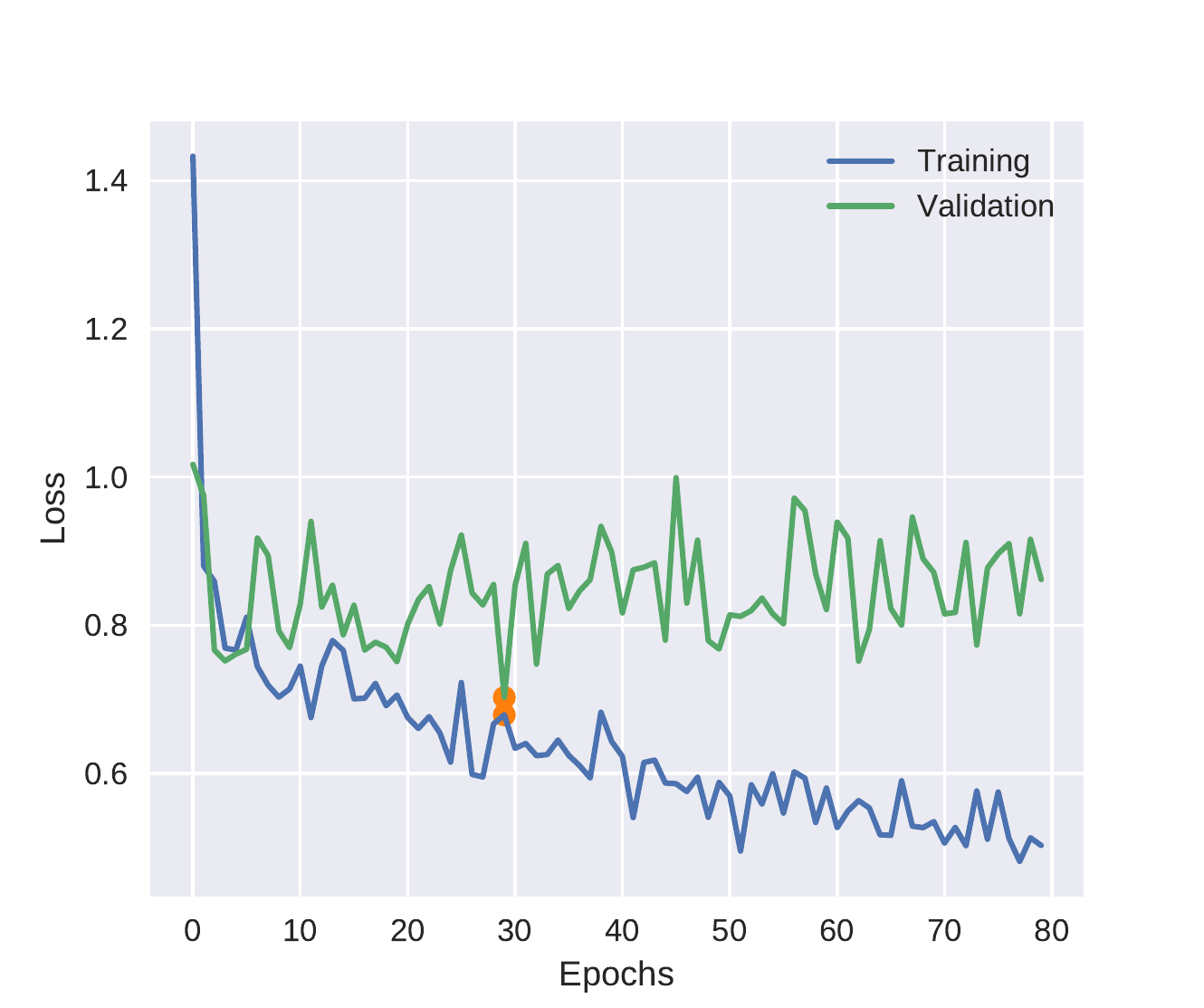} 
	\caption{Training and validation loss of the model~\cite{he2017mask} during the training stage.}
	\label{fig:losses}
\end{figure}

\begin{figure}[h]
\centering
\begin{tabular}{ccc}
	\includegraphics[width=0.23\textwidth]{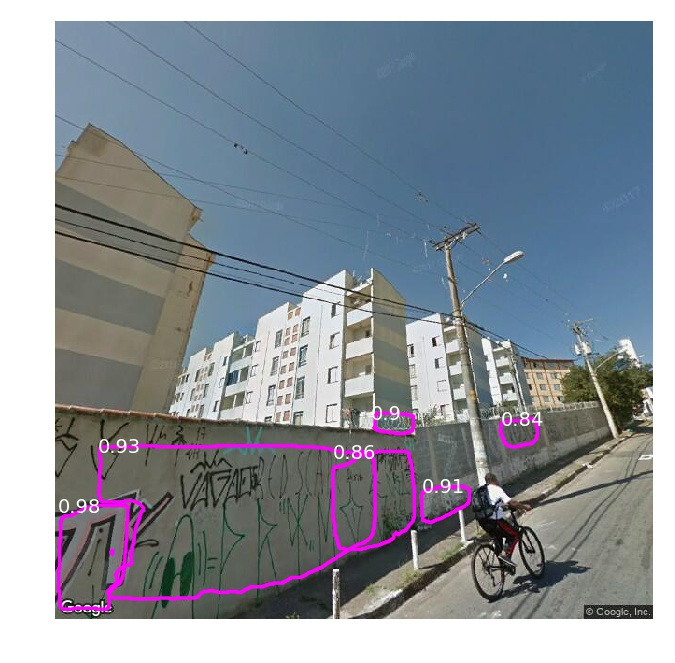} &
	\includegraphics[width=0.23\textwidth]{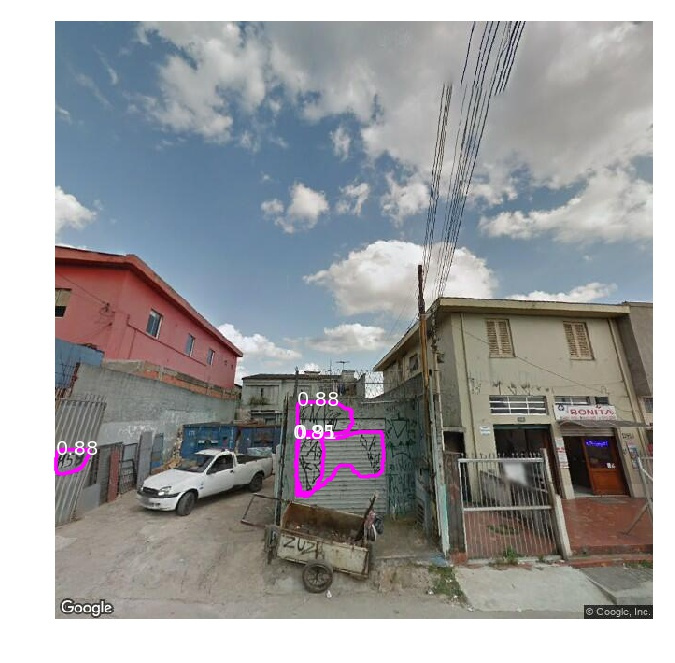} 
\end{tabular}
\caption{Sample of the graffiti detections. The values depicted represent the confidence of the detections.}
\label{fig:detections}
\end{figure}

We initially collected a pilot sample of 10,000 street view images from~\cite{googlemaps} from São Paulo City and a sample was manually chosen. The regions containing graffiti tags were manually identified. A total 632 of images were manually annotated and used to train Resnet 101-layers backbone~\cite{he2016deep} pre-trained on the Coco dataset~\cite{lin2014microsoft}. We used a learning rate of 0.001 and a momentum of 0.9 and trained for 80 epochs. We used the model obtained in epoch 30, given its highest validation error (see Figure~\ref{fig:losses}). The final model showed an average precision~\cite{everingham2010pascal} of $0.57$. Figure~\ref{fig:detections} presents a sample of the detections evaluated. The full time to process a single image is of $0.69$s on a Geforce GTX 1050.

\begin{figure}[ht]
    \centering
    \subfloat[Coverage]{
        \includegraphics[width=0.15\textwidth]{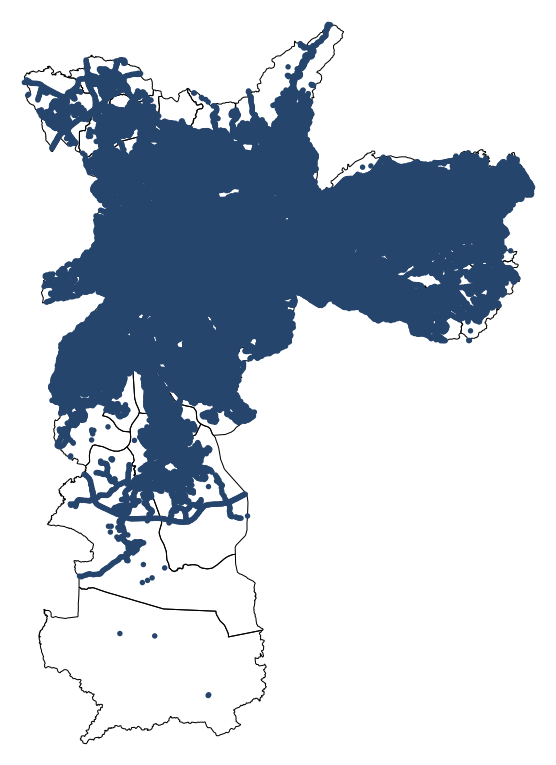} 
 }%
    \qquad
    \subfloat[HDI]{
	\includegraphics[width=0.15\textwidth]{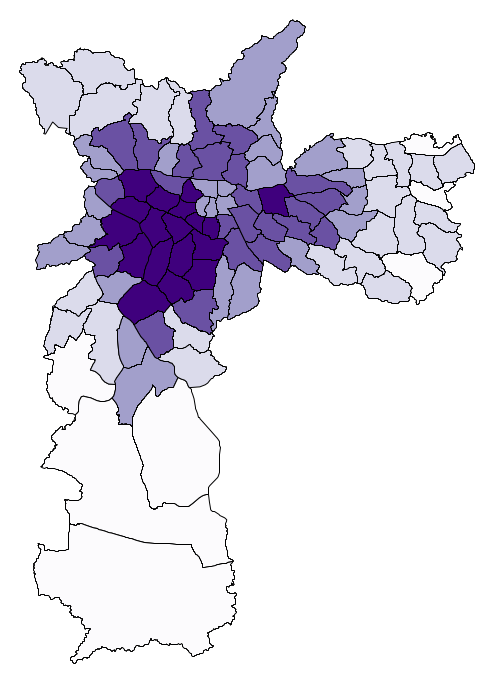} 
    }%
    \caption{Coverage our the utilized images using \cite{googlemaps} images. São Paulo districts 2007 HDI. Data from~\cite{atlas2012idh}.}
    \label{fig:metadata}%
\end{figure}

In Figure~\ref{fig:metadata}~(a) we can see the heterogeneous coverage of the service utilized~\cite{googlemaps} in the city. The two bottommost districts showed little coverage at the time of our acquisition, given the predominantly rural and unpopulated nature of these regions and thus they were not considered in this study.

We used four views for each geographical location, spaced by $90^{\circ}$. Notice in Figure~\ref{fig:views} how the scene elements from the second and the third figures intersect which indicates a full coverage for each geographical location. The majority of the images considered are from 2017 as can be seen in Table~\ref{tab:years}. 

\begin{table}[h]
	\centering
	\setlength\tabcolsep{2pt} % default value: 6pt

	\caption{Acquisition year of the geographical location analyzed. Metadata provided by~\cite{googlemaps}.}
	\begin{tabular}{lccccccccc}
		\toprule
		\textbf{Year}&2010 &2011 &2012 &2013 &2014 &2015 &2016 &2017 &2018 \\
		\midrule

		\textbf{Points}&1,241 &16,311 &207 &422 &2,182 &4,563 &4,211 &39,391 &317 \\
		\bottomrule
	\end{tabular}
	\label{tab:years}
\end{table}
We created a grid over the spatial extent of the city with 134,624 points with 102m vertical and horizontal spacing of our grid. After eliminating images from third-party providers and non-mapped regions (see Figure~\ref{fig:metadata}~(a)), we obtained a geographical coverage of 68,752 geographical points and 275,339 images overall.

\begin{figure}[h]
        \centering
        \includegraphics[width=0.35\textwidth]{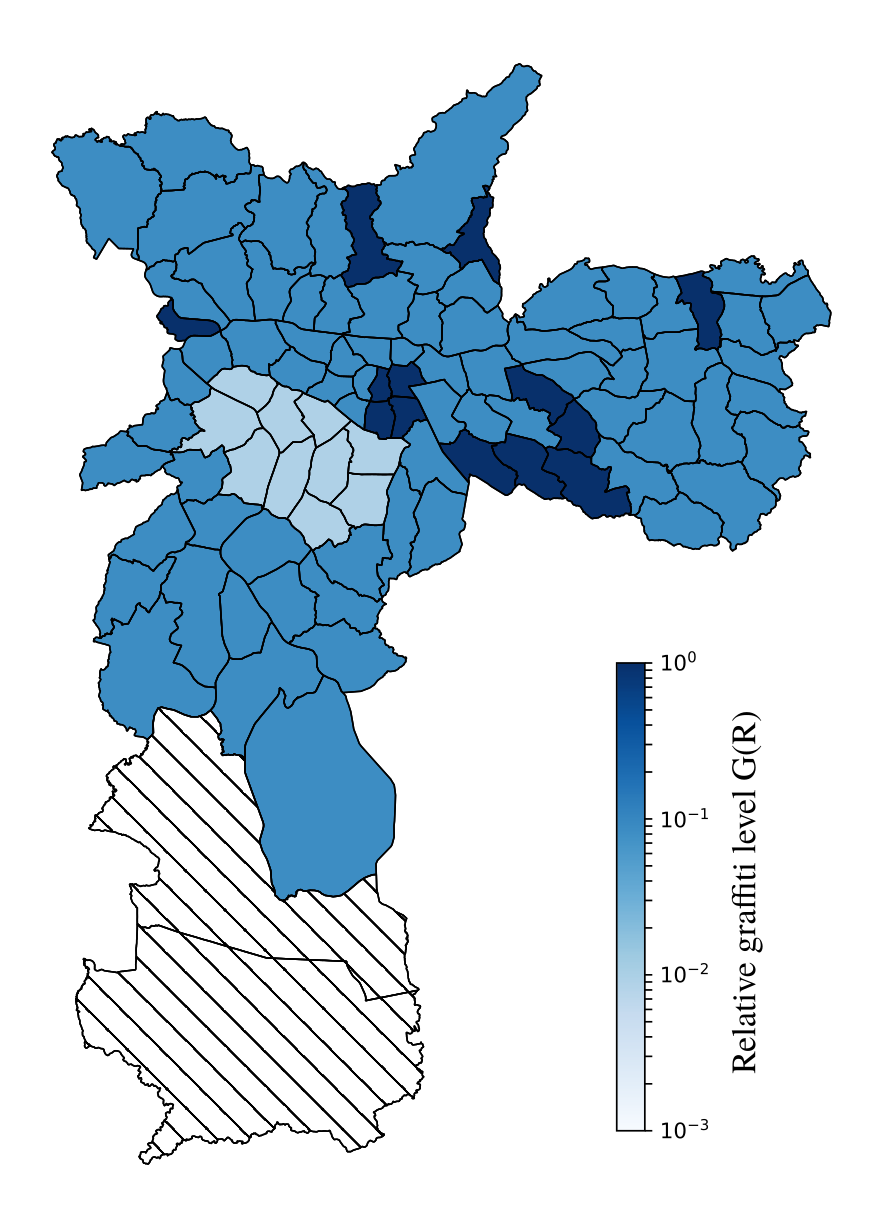} 
	\caption{Relative graffiti level in São Paulo city. The divisions represent the city districts and the colors in logarithmic scale express the relative graffiti level, according to~(\ref{eq:perregion}). Bottommost regions were not considered given the unavailability of data in the region.}
        \label{fig:graffitimap}
\end{figure}

We can see in Figure~\ref{fig:graffitimap} that except a small region inside the map, the regions with highest levels of graffiti are in peripheral regions of the city. The regions with lowest levels of graffiti are in the business center of the city. The bottommost parts were not considered given the coverage of the service utilized.

The Human Development Index (HDI) is a \emph{development} measure of a region that considers life span, income and education aspects~\cite{hdr1990concept}. Figure~\ref{fig:metadata}~(b) is a HDI heat-map by districts using the data released by the city hall~\cite{atlas2012idh} in 2007. Notice that the regions with the lowest levels of graffiti in Figure~\ref{fig:graffitimap} correspond to the regions of highest HDIs in Figure~\ref{fig:metadata}~(b).

%\begin{figure}[ht]
%	\centering
%	\includegraphics[width=0.15\textwidth]{img/idhperdistrict_purple.pdf} 
%	\caption{Human development indices (2007) of São Paulo districts. Data from~\cite{atlas2012idh}.}
%	\label{fig:hdi}
%\end{figure}

%##############################

% ##################### Conclusion
\section{Conclusion}

This work presents an extension of~\cite{tokuda2018identificacao} in the attempt to automaticcaly map regions containing graffiti tags the city.  We systematically collect street view imagery from~\cite{googlemaps} and identify the graffiti tags in each image and we propose a metric for the graffiti level of a geographical region. We did a case study in São Paulo and show that it is in accordance to what is expected given the indicators of HDI.

There are limitations of the proposed approach. One of them is the requirement of sampling, due to computing constraints. Small regions with highly concentrated tagging do not properly contribute to the metric of the region. Ongoing steps include the utilization and the combination of vision algorithms~\cite{lin2018focal,tokuda2013computer} with higher performances and the use of semi-supervised approaches to increase the annotated dataset~\cite{tokuda2018anovel}. Future steps include a denser sampling, a joint analysis with other geographical regions and the use of new datasets that include the same view in different times~\cite{tokuda2018anew}.

%TODO: modify here
The authors thank FAPESP grants \#2014/24918-0, \#2015/22308-2, CNPq, CAPES and NAP eScience - PRP - USP.

\bibliographystyle{IEEEtran}
\bibliography{main}

\end{document}